\documentclass[11pt]{article}

\pdfoutput=1
\usepackage{graphicx,amssymb,amsmath,amsfonts,natbib,fullpage}
\usepackage[obeyspaces,hyphens]{url}
\newcommand{\g}{\,|\,}
\newcommand{\myeq}[1]{Equation~\ref{eq:#1}}

\title{A Tutorial on Bayesian Nonparametric Models}
\author{Samuel J. Gershman$^1$ and David M. Blei$^2$ \\
$^1$Department of Psychology and Neuroscience Institute, Princeton University \\
$^2$Department of Computer Science, Princeton University}

\begin{document}

\maketitle

\begin{abstract}

  A key problem in statistical modeling is model selection, how to
  choose a model at an appropriate level of complexity.  This problem
  appears in many settings, most prominently in choosing the number of
  clusters in mixture models or the number of factors in factor
  analysis.  In this tutorial we describe Bayesian nonparametric
  methods, a class of methods that side-steps this issue by allowing
  the data to determine the complexity of the model.  This tutorial is
  a high-level introduction to Bayesian nonparametric methods and
  contains several examples of their application.

\end{abstract}

\section{Introduction}

How many classes should I use in my mixture model? How many factors
should I use in factor analysis? These questions regularly exercise
scientists as they explore their data.  Most scientists address them
by first fitting several models, with different numbers of clusters or
factors, and then selecting one using model comparison metrics
\citep{claeskens08}.  Model selection metrics usually include two
terms.  The first term measures how well the model fits the data.  The
second term, a complexity penalty, favors simpler models (i.e., ones
with fewer components or factors).

\textit{Bayesian nonparametric (BNP)} models provide a different
approach to this problem~\citep{hjort10}.  Rather than comparing
models that vary in complexity, the BNP approach is to fit a single
model that can adapt its complexity to the data.  Furthermore, BNP
models allow the complexity to grow as more data are observed, such as
when using a model to perform prediction.  For example, consider the
problem of clustering data.  The traditional mixture modeling approach
to clustering requires the number of clusters to be specified in
advance of analyzing the data. The Bayesian nonparametric approach
estimates how many clusters are needed to model the observed data and
allows future data to exhibit previously unseen clusters.\footnote{The
  origins of these methods are in the distribution of random measures
  called the \textit{Dirichlet
    process}~\citep{ferguson73,Antoniak:1974}, which was developed
  mainly for mathematical interest.  These models were dubbed
  ``Bayesian nonparametric'' because they place a prior on the
  infinite-dimensional space of random measures.  With the maturity of
  Markov chain Monte Carlo sampling methods, nearly twenty years
  later, Dirichlet processes became a practical statistical tool
  ~\citep{escobar95}.  Bayesian nonparametric modeling is enjoying a
  renaissance in statistics and machine learning; we focus here on
  their application to latent component models, which is one of their
  central applications.  We describe their formal mathematical
  foundations in Appendix A.}

Using BNP models to analyze data follows the blueprint for Bayesian
data analysis in general~\citep{gelman04}.  Each model expresses a
\emph{generative process} of the data that includes hidden variables.
This process articulates the statistical assumptions that the model
makes, and also specifies the joint probability distribution of the
hidden and observed random variables.  Given an observed data set,
data analysis is performed by \emph{posterior inference}, computing
the conditional distribution of the hidden variables given the
observed data.  Loosely, posterior inference is akin to ``reversing''
the generative process to find the distribution of the hidden
structure that likely generated the observed data.  What distinguishes
Bayesian nonparametric models from other Bayesian models is that the
hidden structure is assumed to grow with the data.  Its complexity,
e.g., the number of mixture components or the number of factors, is
part of the posterior distribution.  Rather than needing to be
specified in advance, it is determined as part of analyzing the data.

In this tutorial, we survey Bayesian nonparametric methods.  We focus
on Bayesian nonparametric extensions of two common models, mixture
models and latent factor models. As we mentioned above, traditional mixture models group
data into a prespecified number of latent clusters.  The Bayesian
nonparametric mixture model, which is called a Chinese restaurant
process mixture (or a Dirichlet process mixture), infers the number of
clusters from the data and allows the number of clusters to grow as new
data points are observed.

Latent factor models decompose observed data into a linear combination
of latent factors.  Different assumptions about the distribution of
factors lead to variants such as factor analysis, principal components
analysis, independent components analysis, and others.  As for
mixtures, a limitation of latent factor models is that the number of
factors must be specified in advance.  The Indian Buffet Process
latent factor model (or Beta process latent factor model) infers the
number of factors from the data and allows the number of factors to grow
as new data points are observed.

We focus on these two types of models because they have served as the
basis for a flexible suite of BNP models.  Models
that are built on BNP mixtures or latent factor models include those
tailored for sequential data \citep{beal02,paisley09b,fox08,fox09},
grouped data \citep{teh06,navarro06}, data in a tree
\citep{johnson07,liang07}, relational data
\citep{kemp06,navarro08,miller09}, and spatial data
\citep{gelfand05,duan07,sudderth09}.

This tutorial is organized as follows. In Sections \ref{sec:mix} and
\ref{sec:ibp} we describe mixture and latent factor models in more
detail, starting from finite-capacity versions and then extending
these to their infinite-capacity counterparts. In Section
\ref{sec:inference} we summarize the standard algorithms for inference
in mixture and latent factor models. Finally, in Section
\ref{sec:limits} we describe several limitations and extensions of
these models. In Appendix A, we detail some of the mathematical and
statistical foundations of BNP models.

We hope to demonstrate how Bayesian nonparametric data analysis
provides a flexible alternative to traditional Bayesian (and
non-Bayesian) modeling.  We give examples of BNP analysis of published
psychological studies, and we point the reader to available software
for performing her own analyses.

\section{Mixture models and clustering}
\label{sec:mix}

In a mixture model, each observed
data point is assumed to belong to a cluster.  In posterior inference,
we infer a grouping or clustering of the data under these
assumptions---this amounts to inferring both the identities of the
clusters and the assignments of the data to them.  Mixture models are
used for understanding the group structure of a data set and for
flexibly estimating the distribution of a population.

For concreteness, consider the problem of modeling response time (RT)
distributions. Psychologists believe that several cognitive processes
contribute to producing behavioral responses \citep{luce86}, and
therefore it is a scientifically relevant question how to decompose
observed RTs into their underlying components. The generative model we
describe below expresses one possible process by which latent causes
(e.g., cognitive processes) might give rise to observed data (e.g.,
RTs).\footnote{A number of papers in the psychology literature have adopted a mixture model approach to modeling RTs \citep[e.g.,][]{ratcliff02,wagenmakers08}. It is worth noting that the decomposition of RTs into constituent cognitive processes performed by the mixture model is fundamentally different from the diffusion model analysis \citep{ratcliff98}, which has become the gold standard in psychology and neuroscience. In the diffusion model, behavioral effects are explained in terms of variations in the underlying parameters of the model, whereas the mixture model attempts to explain these effects in terms of different latent causes governing each response.} Using Bayes' rule, we can invert the generative model to recover
a distribution over the possible set of latent causes of our
observations. The inferred latent causes are commonly known as
``clusters.''

\subsection{Finite mixture modeling}

One approach to this problem is finite mixture modeling.  A finite
mixture model assumes that there are $K$ clusters, each associated
with a parameter $\theta_k$.  Each observation $y_n$ is assumed to be
generated by first choosing a cluster $c_n$ according to $P(c_n)$ and
then generating the observation from its corresponding observation
distribution parameterized by $\theta_{c_n}$.  In the RT modeling
problem, each observation is a scalar RT and each cluster specifies a
hypothetical distribution $F(y_n|\theta_{c_n})$ over the observed
RT.\footnote{The interpretation of a cluster as a psychological
  process must be made with caution.  In our example, the hypothesis
  is that some number of cognitive processes produces the RT data, and
  the mixture model provides a characterization of the cognitive
  process under that hypothesis.  Further scientific experimentation
  is required to validate the existence of these processes and their
  causal relationship to behavior.}

\begin{figure}
\centering
\includegraphics[width=0.6\textwidth]{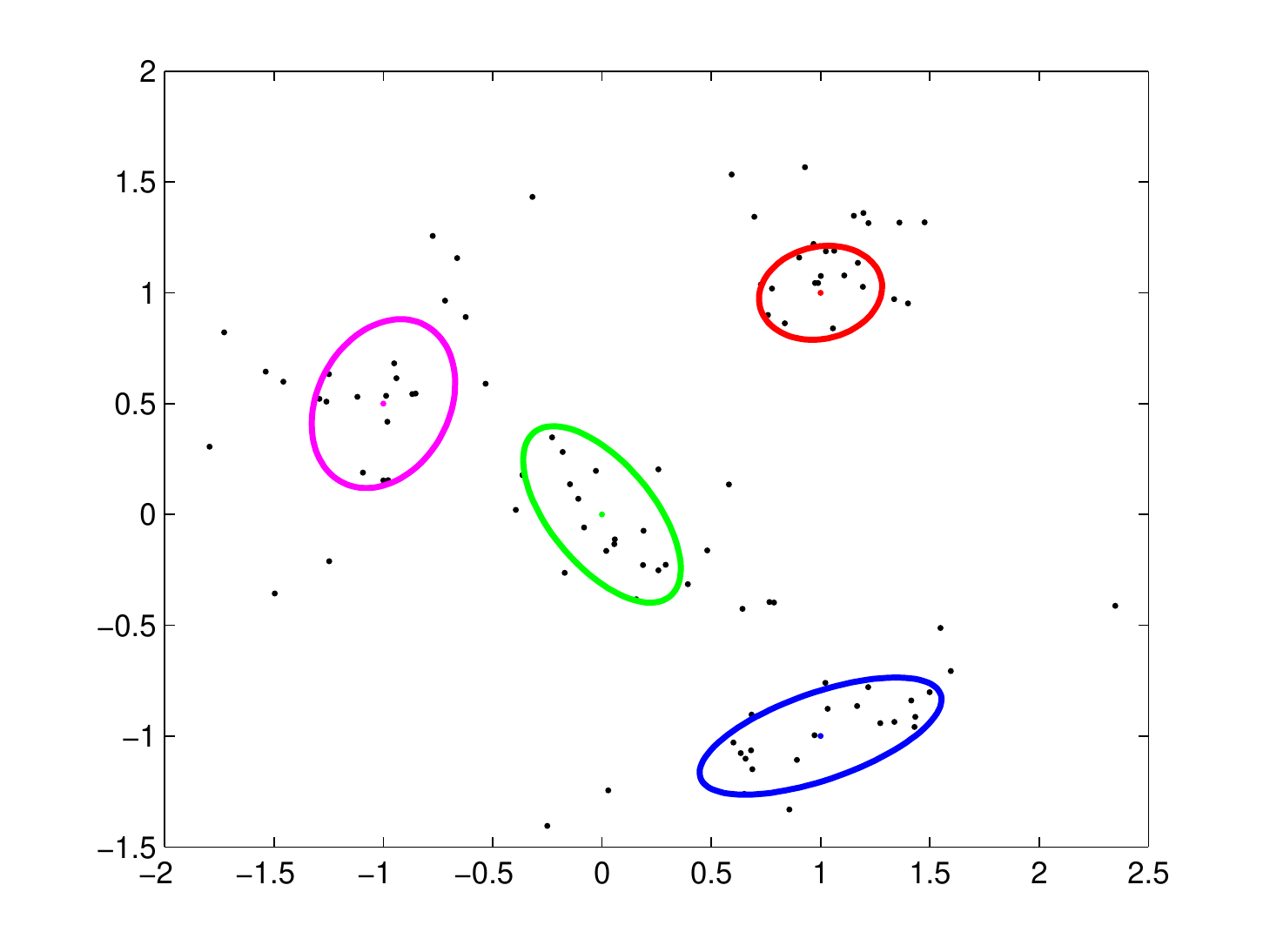}
\caption{\textbf{Draws from a Gaussian mixture model}. Ellipses show
  the standard deviation contour for each mixture component.}
\label{fig:gmm}
\end{figure}

Finite mixtures can accommodate many kinds of data by changing the
data generating distribution.  For example, in a Gaussian mixture
model the data---conditioned on knowing their cluster
assignments---are assumed to be drawn from a Gaussian distribution.  The
cluster parameters $\theta_k$ are the means of the components
(assuming known variances).  Figure~\ref{fig:gmm} illustrates data
drawn from a Gaussian mixture with four clusters.

Bayesian mixture models further contain a prior over the mixing
distribution $P(c)$, and a prior over the cluster parameters: $\theta
\sim G_0$.  (We denote the prior over cluster parameters $G_0$ to
later make a connection to BNP mixture models.)  In
a Gaussian mixture, for example, it is computationally convenient to
choose the cluster parameter prior to be Gaussian.  A convenient
choice for the distribution on the mixing distribution is a Dirichlet.
We will build on fully Bayesian mixture modeling when we discuss
Bayesian nonparametric mixture models.

This generative process defines a joint distribution over the
observations, cluster assignments, and cluster parameters,
\begin{align}
  \label{eq:mixture}
  P(\mathbf{y},\mathbf{c},\theta) = \prod_{k=1}^K G_0(\theta_k)
  \prod_{n=1}^N F(y_n|\theta_{c_n})P(c_n),
\end{align}
where the observations are $\mathbf{y} = \lbrace y_1,\ldots,y_N
\rbrace$, the cluster assignments are $\mathbf{c} = \lbrace
c_1,\ldots,c_N \rbrace$, and the cluster parameters are $\theta =
\lbrace \theta_1,\ldots,\theta_K \rbrace$. The product over $n$
follows from assuming that each observation is conditionally
independent given its latent cluster assignment and the cluster
parameters.  Returning to the RT example, the RTs are assumed to be
independent of each other once we know which cluster generated each RT
and the parameters of the latent clusters.

Given a data set, we are usually interested in the cluster
assignments, i.e., a grouping of the data.\footnote{Under the Dirichlet prior, the
assignment vector $\mathbf{c} = [1,2,2]$ has the same
probability as $\mathbf{c} = [2,1,1]$.  That is, these vectors are
equivalent up to a ``label switch.'' Generally we do not care about
what particular labels are associated with each class; rather, we care about
\emph{partitions}---equivalence classes of assignment vectors that
preserve the same groupings but ignore labels.}  We can use Bayes' rule to
calculate the posterior probability of assignments given the data:
\begin{align}
  P(\mathbf{c}|\mathbf{y}) =
  \frac{P(\mathbf{y}|\mathbf{c})P(\mathbf{c})}{\sum_{\mathbf{c}}
    P(\mathbf{y}|\mathbf{c})P(\mathbf{c})},
\end{align}
where the likelihood is obtained by marginalizing over settings of
$\theta$:
\begin{align}
  P(\mathbf{y}|\mathbf{c}) = \int_{\theta} \left[ \prod_{n=1}^N
    F(\mathbf{y}|\theta_{c_n})\prod_{k=1}^K G_0(\theta_k) \right]
  d\theta.
\end{align}
A $G_0$ that is conjugate to $F$ allows this integral to be calculated
analytically.  For example, the Gaussian is the conjugate prior to a
Gaussian with fixed variance, and this is why it is computationally
convenient to select $G_0$ to be Gaussian in a mixture of Gaussians
model.

The posterior over assignments is intractable because computing the
denominator (marginal likelihood) requires summing over every possible
partition of the data into $K$ groups.  (This problem becomes more
salient in the next section, where we consider the limiting case $K
\rightarrow \infty$.)  We can use approximate methods, such as Markov
chain Monte Carlo \citep{mclachlan00} or variational inference \citep{attias00}; these methods are discussed further in Section \ref{sec:inference}.

\subsection{The Chinese restaurant process}
\label{sec:crpmixture}

When we analyze data with the finite mixture of
Equation~\ref{eq:mixture}, we must specify the number of latent
clusters (e.g., hypothetical cognitive processes) in advance. In many
data analysis settings, however, we do not know this number and would
like to learn it from the data. BNP clustering
addresses this problem by
assuming that there is an infinite number of latent clusters, but that a
finite number of them is used to generate the observed data.  Under
these assumptions, the posterior provides a distribution over the
number of clusters, the assignment of data to clusters, and the
parameters associated with each cluster.  Furthermore, the predictive
distribution, i.e., the distribution of the next data point,
allows for new data to be assigned to a previously unseen cluster.

\begin{figure}
\centering
\includegraphics[width=1.0\textwidth]{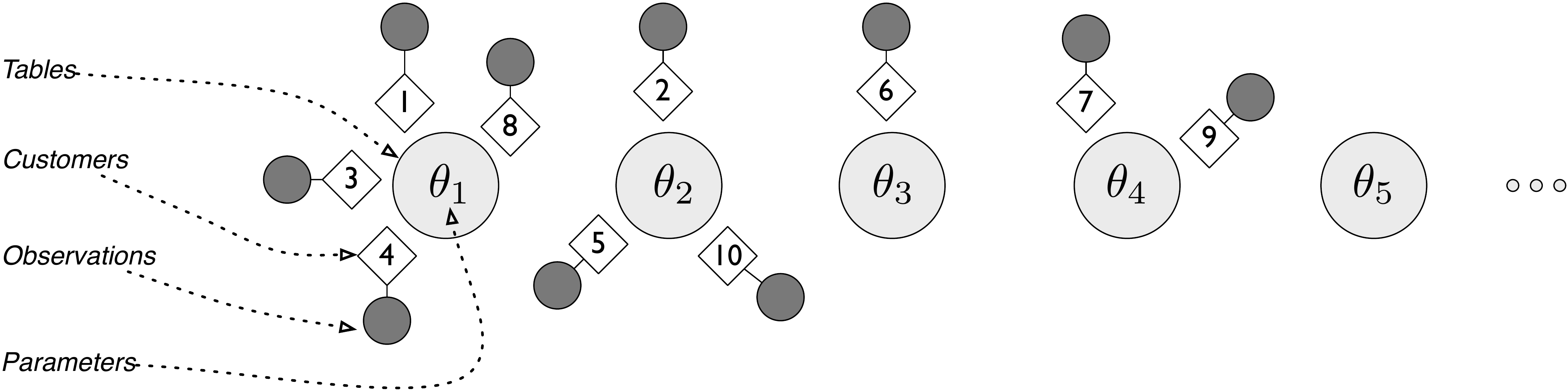}
\caption{\textbf{The Chinese restaurant process}. The generative process of the CRP, where
  numbered diamonds represent customers, attached to their corresponding observations (shaded circles). The large circles
  represent tables (clusters) in the
  CRP and their associated parameters ($\theta$). Note that technically the parameter values $\lbrace \theta \rbrace$ are not part of the CRP \emph{per se}, but rather belong to the full mixture model.}
\label{fig:crp}
\end{figure}

The BNP approach finesses the problem of choosing
the number of clusters by assuming that it is infinite, while
specifying the prior over infinite groupings $P(\mathbf{c})$ in such a
way that it favors assigning data to a small number of groups.  The
prior over groupings is called the \emph{Chinese Restaurant Process} \citep[CRP;][]{aldous85,pitman02}, a distribution over infinite partitions of the
integers; this distribution was independently discovered by \citet{anderson91} in the context of his rational model of categorization (see Section \ref{sec:cog} for more discussion of psychological implications).  The CRP derives its name from the following metaphor.
Imagine a restaurant with an infinite number of tables,\footnote{The
  Chinese restaurant metaphor is due to Pitman and Dubins, who were
  inspired by the seemingly infinite seating capacity of Chinese
  restaurants in San Francisco.} and imagine a sequence of customers
entering the restaurant and sitting down.  The first customer enters
and sits at the first table.  The second customer enters and sits at
the first table with probability $\frac{1}{1+\alpha}$, and the second
table with probability $\frac{\alpha}{1+\alpha}$, where $\alpha$ is a
positive real.  When the $n$th customer enters the restaurant, she
sits at each of the occupied tables with probability proportional to
the number of previous customers sitting there, and at the next
unoccupied table with probability proportional to $\alpha$.  At any
point in this process, the assignment of customers to tables defines a
random partition. A schematic of this process is shown in Figure
\ref{fig:crp}.

More formally, let $c_n$ be the table assignment of the $n$th
customer.  A draw from this distribution can be generated by
sequentially assigning observations to classes with probability
\begin{align}
  P(c_n = k|\mathbf{c}_{1:n-1}) \propto \left\{ \begin{array}{ll}
      \frac{m_k}{n - 1 + \alpha} & \mbox{if $k \leq K_+$ (i.e., $k$ is a previously occupied table)}\\
      \frac{\alpha}{n - 1 + \alpha} & \mbox{otherwise (i.e., $k$ is
        the next unoccupied table)}\end{array} \right.
  \label{eq:crp}
\end{align}
where $m_k$ is the number of customers sitting at table $k$, and $K_+$
is the number of tables for which $m_k > 0$.  The parameter $\alpha$ is called the
\emph{concentration parameter}.  Intuitively, a larger value of
$\alpha$ will produce more occupied tables (and fewer customers per
table).


The CRP exhibits an important invariance property: The cluster
assignments under this distribution are \textit{exchangeable}. This
means that $p(\mathbf{c})$ is unchanged if the order of customers is
shuffled (up to label changes).  This may seem counter-intuitive at
first, since the process in \myeq{crp} is described sequentially.

Consider the joint distribution of a set of customer assignments
$c_{1:N}$.  It decomposes according to the chain rule,
\begin{equation}
  \label{eq:joint}
  p(c_1, c_2, \ldots, c_N) =
  p(c_1) p(c_2 \g c_1) p(c_3 \g c_1, c_2)
  \cdots p(c_N \g c_1, c_2, \ldots, c_{N-1}),
\end{equation}
where each terms comes from \myeq{crp}.  To show that this
distribution is exchangeable, we will introduce some new notation.
Let $K(c_{1:N})$ be the number of groups in which these assignments
place the customers, which is a number between $1$ and $N$.  (Below,
we'll suppress its dependence on $c_{1:N}$.)  Let $I_k$ be the set of
indices of customers assigned to the $k$th group, and let $N_k$ be the
number of customers assigned to that group (i.e., the cardinality of
$I_k$).

Now, examine the product of terms in \myeq{joint} that correspond to
the customers in group $k$.  This product is
\begin{equation}
  \label{eq:pergroup}
  \frac{\alpha \cdot 1 \cdot 2 \cdots (N_k-1)}
  {(I_{k,1} -1 + \alpha) (I_{k,2} - 1 + \alpha)
    \cdots (I_{k,N} - 1 + \alpha)}.
\end{equation}
To see this, notice that the first customer in group $k$ contributes
probability $\frac{\alpha}{I_{k,1} - 1 + \alpha}$ because he is
starting a new table; the second customer contributes probability
$\frac{1}{I_{k,2} - 1 + \alpha}$ because he is sitting a table with
one customer at it; the third customer contributes probability
$\frac{2}{I_{k,3} - 1 + \alpha}$, and so on.  The numerator of
\myeq{pergroup} can be more succinctly written as $\alpha (N_k - 1)!$

With this expression, we now rewrite the joint distribution in
\myeq{joint} as a product over per-group terms,
\begin{equation}
  p(c_{1:N}) = \prod_{k=1}^{K} \frac{\alpha (N_k - 1)!}
  {(I_{k,1} - 1 + \alpha) (I_{k,2} - 1 + \alpha) \cdots
    (I_{k,N_k} - 1 + \alpha)}.
\end{equation}
Finally, notice that the union of $I_k$ across all groups $k$
identifies each index once, because each customer is assigned to
exactly one group.  This simplifies the denominator and lets us write
the joint as
\begin{equation}
  \label{eq:simple-joint}
  p(c_{1:N}) = \frac{\alpha^{K} \prod_{k=1}^{K} (N_k
    - 1)!}{\prod_{i=1}^{N} (i - 1 + \alpha)}.
\end{equation}
\myeq{simple-joint} reveals that \myeq{joint} is exchangeable.  It
only depends on the number of groups $K$ and the size of each group
$N_k$.  The probability of a particular seating configuration
$c_{1:N}$ does not depend on the order in which the customers arrived.

\subsection{Chinese restaurant process mixture models}

The BNP clustering model uses the CRP in an infinite-capacity mixture model
\citep{Antoniak:1974,anderson91,escobar95,rasmussen00}.  Each table $k$ is
associated with a cluster and with a cluster parameter $\theta_k$,
drawn from a prior $G_0$.  We emphasize that there are an infinite
number of clusters, though a finite data set only exhibits a finite
number of active clusters.  Each data point is a ``customer,'' who
sits at a table $c_n$ and then draws its observed value from the
distribution $F(y_n|\theta_{c_n})$. The concentration parameter
$\alpha$ controls the prior expected number of clusters (i.e.,
occupied tables) $K_+$. In particular, this number grows
logarithmically with the number of customers $N$: $\mathbb{E}[K_+] =
\alpha \log N$ (for $\alpha < N/\log N$). If $\alpha$ is treated as unknown, one can put a hyperprior over it and use the same Bayesian machinery discussed in Section \ref{sec:inference} to infer its value.

\begin{figure}
\centering
\includegraphics[width=0.8\textwidth]{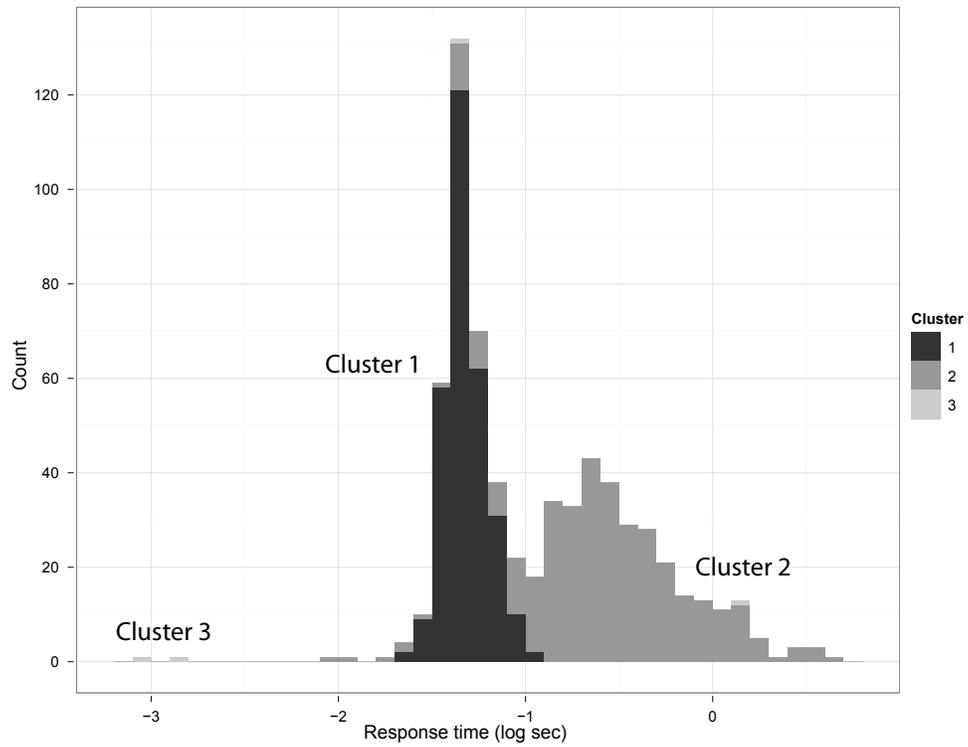}
\caption{\textbf{Response time modeling with the CRP mixture model}. An example distribution of response times from a two-alternative forced-choice decision making experiment \citep{simen09} Colors denote clusters inferred by 100 iterations of Gibbs sampling.}
\label{fig:rt_dp}
\end{figure}

Returning to the RT example, the CRP allows us to place a prior over
partitions of RTs into the hypothetical cognitive processes that
generated them, without committing in advance to a particular number
of processes. As in the finite setting, each process $k$ is associated
with a set of parameters $\theta_k$ specifying the distribution over
RTs (e.g., the mean of a Gaussian for log-transformed RTs). Figure
\ref{fig:rt_dp} shows the clustering of RTs obtained by approximating
the posterior of the CRP mixture model using Gibbs sampling (see
Section \ref{sec:inference}); in this figure, the cluster assignments
from a single sample are shown. These data were collected in an
experiment on two-alternative forced-choice decision making
\citep{simen09}. Notice that the model captures the two primary modes
of the data, as well as a small number of left-skewed outliers.

By examining the posterior over partitions, we can infer both the
assignment of RTs to hypothetical cognitive processes and the number of
hypothetical processes. In addition, the (approximate) posterior provides a measure of confidence in any particular clustering, without committing to a single cluster assignment.
Notice that the number of clusters can grow as more data are observed.
This is both a natural regime for many scientific applications, and it
makes the CRP mixture robust to new data that is far away from the
original observations.

When we analyze data with a CRP, we form an approximation of the
joint posterior over all latent variables and parameters.  In practice, there are two uses for this posterior.  One
is to examine the likely partitioning of the data.  This gives us a
sense of how are data are grouped, and how many groups the CRP model
chose to use.  The second use is to form predictions with the
posterior predictive distribution.  With a CRP mixture, the posterior predictive
distribution is
\begin{equation}
P(y_{n+1}|\mathbf{y}_{1:n}) = \sum_{\mathbf{c}_{1:n+1}} \int_{\theta} P(y_{n+1}|c_{n+1},\theta) P(c_{n+1}|\mathbf{c}_{1:n}) P(\mathbf{c}_{1:n},\theta|\mathbf{y}_{1:n}) d\theta.
\end{equation}
Since the CRP prior, $P(c_{n+1}|\mathbf{c}_{1:n})$, appears in the predictive distribution, the CRP mixture allows new data to possibly
exhibit a previously unseen cluster.

\section{Latent factor models and dimensionality reduction}
\label{sec:ibp}

Mixture models assume that each observation is assigned to one of $K$
components.  Latent factor models weaken this assumption: each
observation is influenced by each of $K$ components in a different way
\citep[see][for an overview]{comrey92}. These models have a long
history in psychology and psychometrics \citep{pearson01,thurstone31},
and one of their first applications was to modeling human intelligence
\citep{spearman04}. We will return to this application shortly.

Latent factor models provide dimensionality reduction in the (usual)
case when the number of components is smaller than the dimension of
the data.  Each observation is associated with a vector of component
activations (latent factors) that describes how much each component
contributes to it, and this vector can be seen as a lower dimensional
representation of the observation itself.  When fit to data, the
components parsimoniously capture the primary modes of variation in
the observations.

The most popular of these models---factor analysis (FA), principal
component analysis (PCA) and independent components analysis
(ICA)---all assume that the number of factors ($K$) is
known.  The Bayesian nonparametric variant of latent factor models we describe below
allows the number of factors to grow as more data are observed.  As
with the BNP mixture model, the posterior
distribution provides both the properties of the latent factors and
how many are exhibited in the data.\footnote{Historically, psychologists have explored a variety of rotation methods for enforcing sparsity and interpretability in FA solutions, starting with early work summarized by \citet{thurstone47}. Many recent methods are reviewed by \citet{browne01}. The Bayesian approach we adopt differs from these methods by specifying a preference for certain kinds of solutions in terms of the prior.}

In classical factor analysis, the observed data is a collection of $N$ vectors, $\mathbf{Y} = \lbrace
\mathbf{y}_1,\ldots,\mathbf{y}_N \rbrace$, each of which are $M$-dimensional. Thus, $\mathbf{Y}$ is a matrix where rows correspond to observations and columns correspond to observed dimensions. The data (e.g., intelligence test scores) are assumed to be
generated by a noisy weighted combination of latent factors (e.g., underlying intelligence faculties):
\begin{align}
\mathbf{y}_n = \mathbf{G} \mathbf{x}_n + \boldsymbol\epsilon_n,
\label{eq:fa}
\end{align}
where $\mathbf{G}$ is a $M \times K$ factor loading matrix expressing how latent factor $k$ influences
observation dimension $m$, $\mathbf{x}_n$ is a $K$-dimensional vector expressing the activity of each latent factor, and $\boldsymbol{\epsilon}_n$ is a vector of independent Gaussian noise terms.\footnote{The
assumption of Gaussian noise in Eq. \ref{eq:fa} is not fundamental to the latent factor model, but
is the most common choice of noise distribution.}  We can extend this to a
sparse model by decomposing the factor loading into the product of two components: $G_{mk} = z_{mk} w_{mk}$, where $z_{mk}$ is a binary ``mask'' variable that indicates whether factor $k$ is ``on'' ($z_{mk}=1$) or ``off'' ($z_{mk}=0$) for dimension $m$, and $w_{mk}$ is a continuous weight variable. This is sometimes called a ``spike and slab'' model \citep{mitchell88,ishwaran05} because the marginal distribution over $x_{mk}$ is a mixture of a (typically Gaussian) ``slab'' $P(w_{mk})$ over the space of latent factors and a ``spike'' at zero, $P(z_{mk}=0)$.

We take a Bayesian approach to inferring the latent factors, mask
variables, and weights.  We place priors over them and use Bayes' rule
to compute the posterior
$P(\mathbf{G},\mathbf{Z},\mathbf{W}|\mathbf{Y})$. In contrast,
classical techniques like ICA, FA and PCA fit point estimates of the
parameters (typically maximum likelihood estimates).

As mentioned above, a classic application of factor analysis in
psychology is to the modeling of human intelligence
\citep{spearman04}. \citet{spearman04} argued that there exists a
general intelligence factor (the so-called \emph{g}-factor) that can
be extracted by applying classical factor analysis methods to
intelligence test data. Spearman's hypothesis was motivated by the
observation that scores on different tests tend to be correlated:
Participants who score highly on one test are likely to score highly
on another. However, several researchers have disputed the notion that
this pattern arises from a unitary intelligence construct, arguing
that intelligence consists of a multiplicity of components
\citep{gould81}. Although we do not aspire to resolve this
controversy, the question of how many factors underlie human
intelligence is a convenient testbed for the BNP factor
analysis model described below.

Since in reality the number of latent intelligence factors is unknown,
we would like to avoid specifying $K$ and instead allow the data to
determine the number of factors. Following the model proposed by
\citet{knowles11}, $\mathbf{Z}$ is a binary matrix with a finite
number of rows (each corresponding to an intelligence measure) and an
infinite number of columns (each corresponding to a latent factor).

\begin{figure}
\centering
\includegraphics[width=1.0\textwidth]{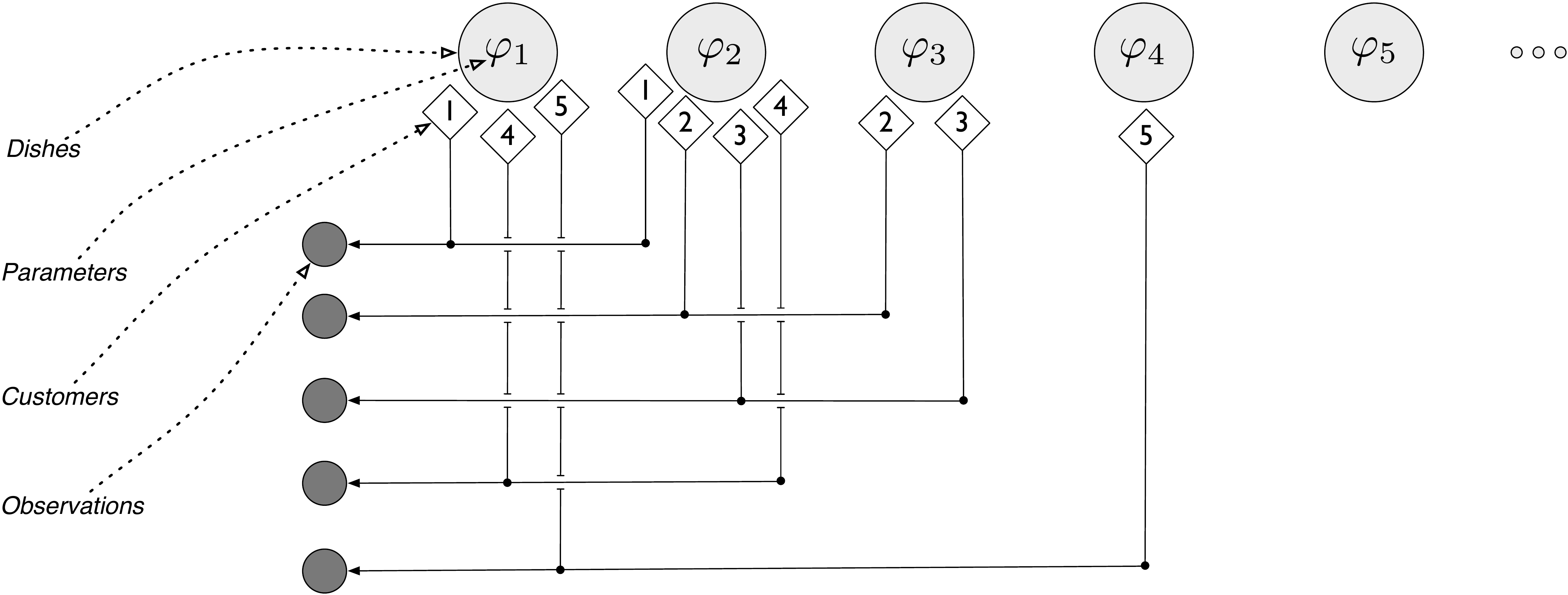}
\caption{\textbf{The Indian buffet process}.
  The generative process of the IBP, where
  numbered diamonds represent customers, attached to their corresponding observations (shaded circles). Large circles
  represent dishes (factors) in the
  IBP, along with their associated parameters ($\varphi$).
Each customer selects several dishes, and each customer's observation (in the latent factor model)
is a linear combination of the selected dish's parameters. Note that technically the parameter values $\lbrace \phi \rbrace$ are not part of the IBP \emph{per se}, but rather belong to the full latent factor model.}
\label{fig:ibp}
\end{figure}

\begin{figure}
\centering
\includegraphics[width=0.8\textwidth]{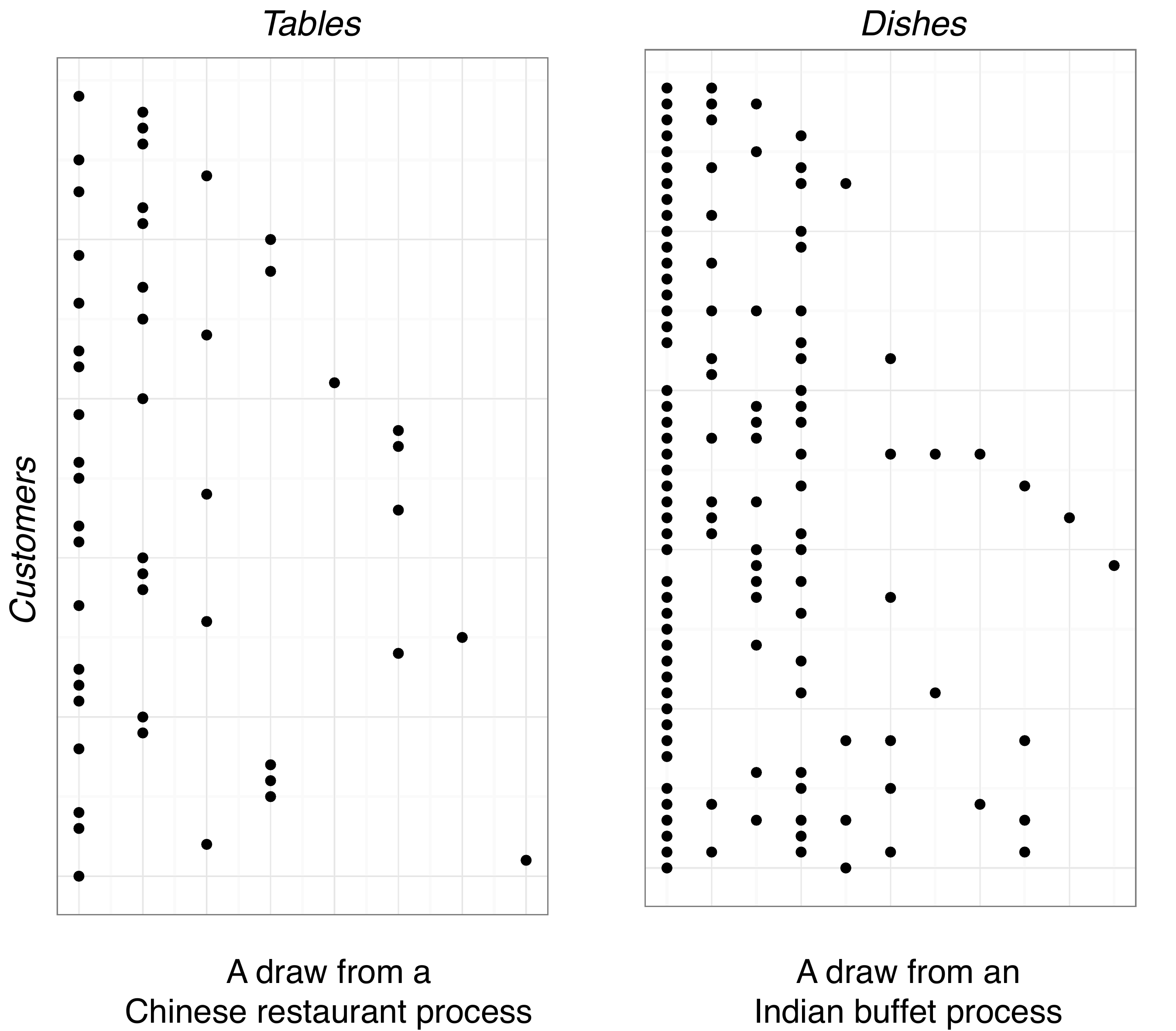}
\caption{\textbf{Draws from the CRP and IBP}. (\emph{Left}) Random
  draw from the Chinese restaurant process. (\emph{Right}) Random draw
  from the Indian buffet process. In the CRP, each customer is
  assigned to a single component.  In the IBP, a customer can be
  assigned to multiple components.}
\label{fig:ibp_draw}
\end{figure}

Like the CRP, the infinite-capacity distribution over $\mathbf{Z}$ has
been furnished with a similarly colorful culinary metaphor, dubbed the
\emph{Indian buffet process} \citep[IBP;][]{griffiths05,griffiths11}. A customer
(dimension) enters a buffet with an infinite number of dishes
(factors) arranged in a line. The probability that a customer $m$ samples
dish $k$ (i.e., sets $z_{mk}=1$) is proportional to its popularity
$h_k$ (the number of prior customers who have sampled the dish). When
the customer has considered all the previously sampled dishes (i.e.,
those for which $h_k>0$), she samples an additional
Poisson($\alpha/N$) dishes that have never been sampled before. When
all $M$ customers have navigated the buffet, the resulting binary
matrix $\mathbf{Z}$ (encoding which customers sampled which dishes) is
a draw from the IBP.

The IBP plays the same role for latent factor models that the CRP
plays for mixture models: It functions as an infinite-capacity prior
over the space of latent variables, allowing an unbounded number of
latent factors \citep{knowles11}. Whereas in the CRP, each observation
is associated with only one latent component, in the IBP each
observation (or, in the factor analysis model described above, each
dimension) is associated with a theoretically infinite number of
latent components.\footnote{Most of these latent factors will be
  ``off'' because the IBP preserves the sparsity of the finite
  Beta-Bernoulli prior \citep{griffiths05}. The degree of sparsity is
  controlled by $\alpha$: for larger values, more latent factors will
  tend to be active.} A schematic of the IBP is shown in Figure
\ref{fig:ibp}. Comparing to Figure \ref{fig:crp},
the key difference between the CRP and the IBP is that in the CRP,
each customer sits at a single table, whereas in the IBP, a customer
can sample several dishes. This difference is illustrated in Figure \ref{fig:ibp_draw}, which shows random draws from both models side-by-side.

Returning to the intelligence modeling example, posterior inference in the
infinite latent factor model yields a distribution over matrices of
latent factors which describe hypothetical intelligence structures:
\begin{align}
P(\mathbf{X},\mathbf{W},\mathbf{Z}|\mathbf{Y}) \propto P(\mathbf{Y}|\mathbf{X},\mathbf{W},\mathbf{Z}) P(\mathbf{X}) P(\mathbf{W}) P(\mathbf{Z}).
\end{align}
Exact inference is intractable because the normalizing constant
requires a sum over all possible binary matrices.  However, we can
approximate the posterior using one of the techniques described in the
next section (e.g., with a set of samples). Given posterior samples of
$\mathbf{Z}$, one typically examines the highest-probability sample
(the \emph{maximum a posteriori}, or MAP, estimate) to get a sense of
the latent factor structure. As with the CRP, if one is interested in
predicting some function of $\mathbf{Z}$, then it is best to average
this function over the samples.

\begin{figure}
\centering
\includegraphics[width=0.8\textwidth]{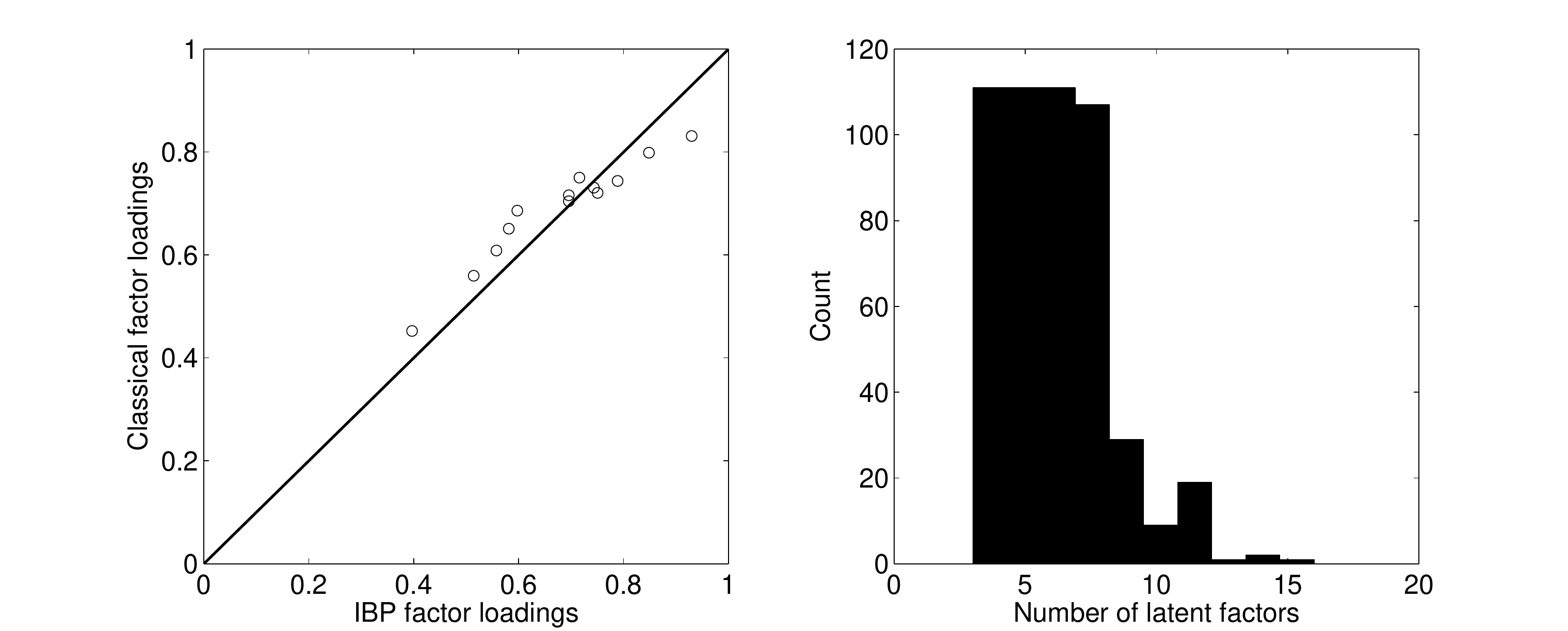}
\caption{\textbf{IBP factor analysis of human performance on reasoning tasks}. (\emph{Left})
  Histogram of the number of latent factors inferred by Gibbs sampling applied to reasoning task data from \citet{kane04}. 1000 samples were generated, and the first 500 were discarded as burn-in. (\emph{Right}) Relationship between the loading of the first factor inferred by IBP factor analysis and Spearman's \emph{g} \citep[i.e., the loading of the first factor inferred by classical factor analysis;][]{spearman04}.}
\label{fig:ibpfa}
\end{figure}

Figure \ref{fig:ibpfa} shows the results of the IBP factor analysis applied to data collected by \citet{kane04}. We consider the 13 reasoning tasks administered to 234 participants. The left panel displays a histogram of the factor counts (the number of times $z_{mk}=1$ across posterior samples). This plot indicates that the dataset is best described by a combination of around $4-7$ factors; although this is obviously not a conclusive argument against the existence of a general intelligence factor, it suggests that additional factors merit further investigation. The right panel displays the first factor loading from the IBP factor analysis plotted against the \emph{g}-factor, demonstrating that the nonparametric method is able to extract structure consistent with classical methods.\footnote{It is worth noting that the field of intelligence research has developed its methods far beyond Spearman's \emph{g}-factor. In particular, \emph{hierarchical} factor analysis is now in common use. See \citet{kane04} for an example.}

\section{Inference}

\label{sec:inference}

We have described two classes of BNP models---mixture models based on
the CRP and latent factor models based on the IBP.  Both types of
models posit a generative probabilistic process of a collection of
observed (and future) data that includes hidden structure.  We analyze
data with these models by examining the posterior distribution of the
hidden structure given the observations; this gives us a distribution
over which latent structure likely generated our data.

Thus, the basic computational problem in BNP modeling (as in most of
Bayesian statistics) is computing the posterior. For many interesting
models---including those discussed here---the posterior is not
available in closed form.  There are several ways to approximate it.
While a comprehensive treatment of inference methods in BNP models is
beyond the scope of this tutorial, we will describe some of the most
widely-used algorithms. In Appendix B, we provide links to software packages implementing these algorithms.

The most widely used posterior inference methods in Bayesian
nonparametric models are Markov Chain Monte Carlo (MCMC) methods.  The
idea MCMC methods is to define a Markov chain on the hidden variables
that has the posterior as its equilibrium
distribution~\citep{andrieu03}.  By drawing samples from this Markov
chain, one eventually obtains samples from the posterior.  A simple
form of MCMC sampling is Gibbs sampling, where the Markov chain is
constructed by considering the conditional distribution of each hidden
variable given the others and the observations.  Thanks to the
exchangeability property described in Section~\ref{sec:crpmixture},
CRP mixtures are particularly amenable to Gibbs sampling---in
considering the conditional distributions, each observation can be
considered to be the ``last'' one and the distribution of
Equation~\ref{eq:crp} can be used as one term of the conditional
distribution.  (The other term is the likelihood of the observations
under each partition.)  \citet{neal00} provides an excellent survey of
Gibbs sampling and other MCMC algorithms for inference in CRP mixture
models \citep[see
also][]{escobar95,rasmussen00,ishwaran01,jain04,fearnhead04,wood07}. Gibbs
sampling for the IBP factor analysis model is described in
\citet{knowles11}.

\begin{figure}
\centering
\includegraphics[width=0.9\textwidth]{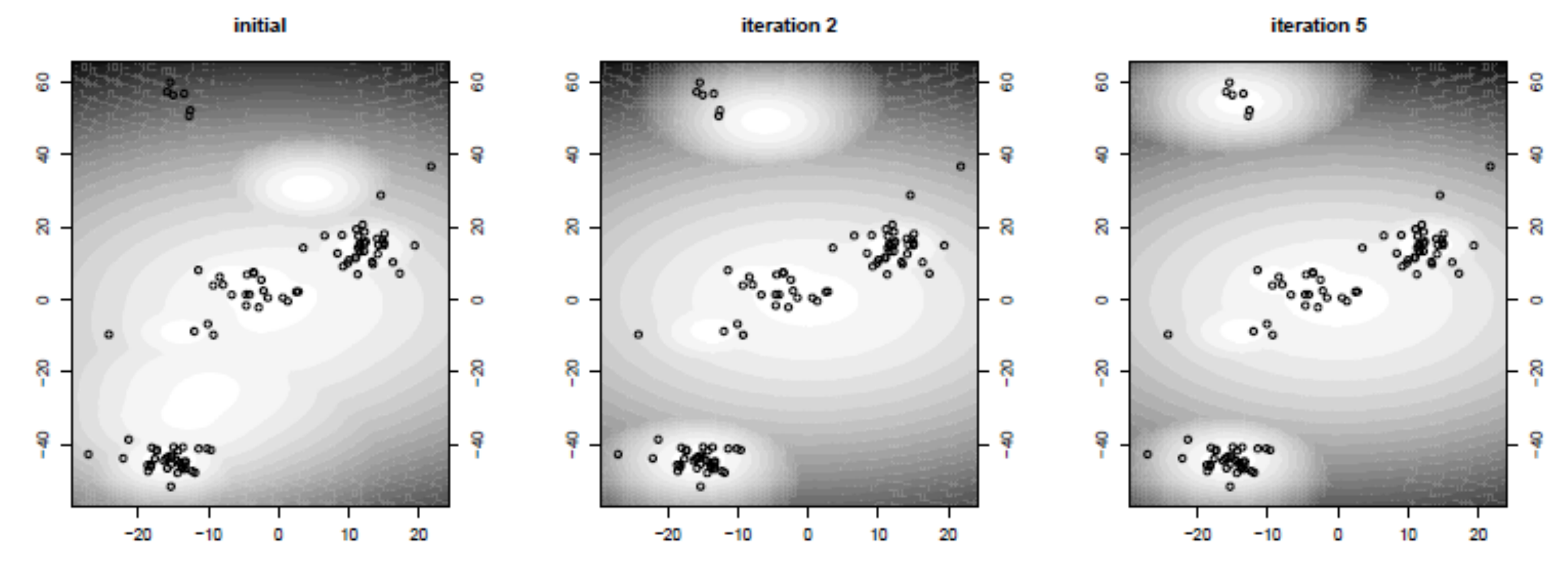}
\caption{\textbf{Inference in a Chinese restaurant process mixture model}. The approximate
  predictive distribution given by variational inference at different
  stages of the algorithm. The data are 100 points generated by a
  Gaussian DP mixture model with fixed diagonal covariance. Figure
  reproduced with permission from \citet{blei06}.}
	\label{fig:dpvb}
\end{figure}

MCMC methods, although guaranteed to converge to the posterior with
enough samples, have two drawbacks: (1) The samplers must be run for
many iterations before convergence and (2) it is difficult to assess
convergence.  An alternative approach to approximating the posterior
is \emph{variational inference} \citep{jordan99}.  This approach is
based on the idea of approximating the posterior with a simpler family
of distributions and searching for the member of that family that is
closest to it.\footnote{Distance between probability distributions in
  this setting is measured by the Kullback-Leibler divergence
  (relative entropy).}  Although variational methods are not
guaranteed to recover the true posterior (unless it belongs to the
simple family of distributions), they are typically faster than MCMC
and convergence assessment is straightforward. These methods have been
applied to CRP mixture models \citep[][ see Fig. \ref{fig:dpvb} for an
example]{blei06,kurihara07} and IBP latent factor models
\citep{doshi09,paisley10}. We note that variational inference
  usually operates on a the random measure representation of CRP
  mixtures and IBP factor models, which are described in Appendix A.
  Gibbs samplers that operate on this representation are also
  available~\citep{ishwaran01}.

As we mentioned in the introduction, BNP methods provide an
alternative to model selection over a parameterized family of
models.\footnote{The Journal of Mathematical Psychology has published
  two special issues \citep{myung00,wagenmakers06} on
  model selection which review a broad array of model selection
  techniques (both Bayesian and non-Bayesian).}  In effect, both MCMC and variational
strategies for posterior inference provide a data-directed mechanism
for simultaneously searching the space of models and finding optimal
parameters.  This is convenient in settings like mixture modeling or
factor analysis because we avoid needing to fit models for each
candidate number of components.  It is essential in more complex
settings, where the algorithm searches over a space that is difficult
to efficiently enumerate and explore.


\section{Limitations and extensions}
\label{sec:limits}

We have described the most widely used BNP models, but this is only
the tip of the iceberg.  In this section we highlight some key
limitations of the models described above, and the extensions that
have been developed to address these limitations. It is worth
mentioning here that we cannot do full justice to the variety of BNP
models that have been developed over the past 40 years; we have
omitted many exciting and widely-used ideas, such as Pitman-Yor
processes, gamma processes, Dirichlet diffusion trees and Kingman's
coalescent. To learn more about these ideas, see the recent volume
edited by \citet{hjort10}.

\subsection{Hierarchical structure}

The first limitation concerns \emph{grouped} data: how can we capture
both commonalities and idiosyncrasies across individuals within a
group? For example, members of an animal species will tend to be
similar to each other, but also unique in certain ways. The standard
Bayesian approach to this problem is based on \emph{hierarchical
  models} \citep{gelman04}, in which individuals are coupled by virtue
of being drawn from the same group-level distribution.\footnote{See also the recent issue of Journal of Mathematical Psychology (Volume 55, Issue 1) devoted to hierarchical Bayesian models. \citet{lee10} provides an overview for cognitive psychologists.} The parameters
of this distribution govern both the characteristics of the group and
the degree of coupling. In the nonparametric setting, hierarchical
extensions of the Dirichlet process \citep{teh06} and beta process
\citep{thibaux07} have been developed, allowing an infinite number of
latent components to be shared by multiple individuals. For example,
hierarchical Dirichlet processes can be applied to modeling text
documents, where each document is represented by an infinite mixture
of word distributions (``topics'') that are shared across documents.

Returning to the RT example from Section \ref{sec:mix}, imagine measuring RTs for several subjects. The goal again is to infer which underlying cognitive process generated each response time. Suppose we assume that the same cognitive processes are shared across subjects, but they may occur in different proportions.  This is precisely the kind of structure the HDP can capture.

\subsection{Time series models}

The second limitation concerns \emph{sequential} data: how can we
capture dependencies between observations arriving in a sequence? One
of the most well-known models for capturing such dependencies is the
hidden Markov model \citep[see, e.g.,][]{bishop06}, in which the latent
class for observation $n$ depends on the latent class for observation
$n-1$. The infinite hidden Markov model
\citep[HMM;][]{beal02,teh06,paisley09b} posits the same sequential
structure, but employs an infinite number of latent
classes. \citet{teh06} showed that the infinite HMM is a special case
of the hierarchical Dirichlet process.

As an alternative to the HMM (which assumes a discrete latent state),
a linear dynamical system (also known as an autoregressive moving average model) assumes that the
latent state is continuous and evolves over time according to a
linear-Gaussian Markov process. In a \emph{switching} linear dynamical
system, the system can have a number of dynamical modes; this allows
the marginal transition distribution to be non-linear. \citet{fox08} have explored nonparametric variants of
switching linear dynamical systems, where the number of dynamical
modes is inferred from the data using an HDP prior.

\subsection{Spatial models}

Another type of dependency arising in many datasets is
\emph{spatial}. For example, one expects that if a disease occurs in
one location, it is also likely to occur in a nearby location. One way
to capture such dependencies in a BNP model is to make the base
distribution $G_0$ of the DP dependent on a location variable
\citep{gelfand05,duan07}. In the field of computer vision, \citet{sudderth09} have applied a spatially-coupled
generalization of the DP to the task of image segmentation, allowing
them to encode a prior bias that nearby pixels belong to the same
segment.

We note in passing a burgeoning area of research attempting to devise
more general specifications of dependencies in BNP models,
particularly for DPs \citep{maceachern99,griffin06,blei10}. These
dependencies could be arbitrary functions defined over a set of
covariates (e.g., age, income, weight). For example, people with
similar age and weight will tend to have similar risks for certain
diseases.

More recently, several authors have attempted to apply these ideas to
the IBP and latent factor models
\citep{miller08,doshi09b,williamson10}.

\subsection{Supervised learning}

We have restricted ourselves to a discussion of \emph{unsupervised}
learning problems, where the goal is to discover hidden structure in
data. In \emph{supervised} learning, the goal is to predict some
output variable given a set of input variables (covariates). When the
output variable is continuous, this corresponds to \emph{regression};
when the output variable is discrete, this corresponds to
\emph{classification}.

For many supervised learning problems, the outputs are non-linear
functions of the inputs. The BNP approach to this problem is to place
a prior distribution (known as a \emph{Gaussian process}) directly
over the space of non-linear functions, rather than specifying a
parametric family of non-linear functions and placing priors over
their parameters. Supervised learning proceeds by posterior inference
over functions using the Gaussian process prior. The output of
inference is itself a Gaussian process, characterized by a mean
function and a covariance function (analogous to a mean vector and
covariance matrix in parametric Gaussian models). Given a new set of
inputs, the posterior Gaussian process induces a predictive
distribution over outputs. Although we do not discuss this approach
further, \cite{rasmussen06} is an excellent textbook on this
topic.

Recently, another nonparametric approach to supervised learning has
been developed, based on the CRP mixture model \citep{shahbaba09,hannah10}. The idea
is to place a DP mixture prior over the inputs and then model the mean
function of the outputs as conditionally linear within each mixture
component \citep[see also][for related approaches]{rasmussen02,meeds06}. The result is a marginally non-linear model of the outputs
with linear sub-structure. Intuitively, each mixture component
isolates a region of the input space and models the mean output
linearly within that region. This is an example of a \emph{generative}
approach to supervised learning, where the joint distribution over
both the inputs and outputs is modeled. In contrast, the Gaussian
process approach described above is a \emph{discriminative} approach,
modeling only the conditional distribution of the outputs given the
inputs.

\section{Conclusions}

BNP models are an emerging set of statistical tools
for building flexible models whose structure grows and adapts to data.
In this tutorial, we have reviewed the basics of BNP modeling and illustrated their potential in scientific
problems.

It is worth noting here that while BNP models address the problem of choosing the number of mixture components or latent factors, they are not a general solution to the model selection problem which has received extensive attention within mathematical psychology and other disciplines \citep[see][for a comprehensive treatment]{claeskens08}. In some cases, it may be preferable to place a prior over finite-capacity models and then compare Bayes factors \citep{kass95,vanpaemel10}, or to use selection criteria motivated by other theoretical frameworks, such as information theory \citep{grunwald07}.

\subsection{Bayesian nonparametric models of cognition}
\label{sec:cog}

We have treated BNP models purely as a data analysis tool. However,
there is a flourishing tradition of work in cognitive psychology on
using BNP models as theories of cognition. The earliest example dates
back to \citet{anderson91}, who argued that a version of the CRP
mixture model could explain human categorization behavior. The idea in this model is that
humans adaptively learn the number of categories from their
observations. A number of recent authors have extended this work
\citep{griffiths07,heller09,sanborn10} and applied it to other
domains, such as classical conditioning \citep{gershman10}.

The IBP has also been applied to human cognition. In particular, \citet{austerweil09} argued that humans decompose visual stimuli into latent features in a manner consistent with the IBP. When the parts that compose objects strongly covary across objects, humans treat whole objects as features, whereas individual parts are treated as features if the covariance is weak. This finding is consistent with the idea that the number of inferred features changes flexibly with the data.

BNP models have been fruitfully applied in several other domains,
including word segmentation \citep{goldwater09}, relational theory
acquisition \citep{kemp10} and function learning \citep{griffiths09}.

\subsection{Suggestions for further reading}

A recent edited volume by \citet{hjort10} is a useful resource on
applied Bayesian nonparametrics. For a more general introduction to
statistical machine learning with probabilistic models, see
\citet{bishop06}.  For a review of applied Bayesian statistics,
see \citet{gelman04}.

\section*{Acknowledgements}

We are grateful to Andrew Conway, Katherine Heller, Irvin Hwang, Ed
Vul, James Pooley and Ken Norman who offered comments on an earlier
version of the manuscript. The manuscript was greatly improved by
suggestions from the reviewers. We are also grateful to Patrick Simen,
Andrew Conway and Michael Kane for sharing their data and offering helpful suggestions. This work was
supported by a graduate research fellowship from
the NSF to SJG.  DMB is supported
by ONR 175-6343, NSF CAREER 0745520, AFOSR 09NL202, the Alfred
P. Sloan foundation, and a grant from Google.

\newpage

\section*{Appendix A: Foundations}

We have developed BNP methods via the CRP and
IBP, both of which are priors over combinatorial
structures (infinite partitions and infinite binary matrices).  These
are the easiest first ways to understand this class of models, but
their mathematical foundations are found in constructions of random
distributions.  In this section, we review this perspective of the CRP
mixture and IBP factor model.


\subsection*{\textbf{The Dirichlet process}}

The Dirichlet process (DP) is a distribution over distributions.  It is
parameterized by a concentration parameter $\alpha > 0$ and a base
distribution $G_0$, which is a distribution over a space $\Theta$.  A
random variable drawn from a DP is itself a
distribution over $\Theta$.  A random distribution $G$ drawn from a DP is denoted
$G \sim \textrm{DP}(\alpha, G_0)$.

The DP was first developed in~\cite{ferguson73}, who showed
its existence by appealing to its finite dimensional distributions.
Consider a measurable partition of $\Theta$, $\lbrace T_1,\ldots,T_K
\rbrace$.\footnote{A partition of $\Theta$ defines a collection of
  subsets whose union is $\Theta$. A partition is measurable if it is
  closed under complementation and countable union.}  If $G \sim
\mbox{DP}(\alpha,G_0)$ then every measurable partition of $\Theta$ is
Dirichlet-distributed,
\begin{align}
  (G(T_1),\ldots,G(T_K)) \sim \mbox{Dir}
  (\alpha G_0(T_1),\ldots,\alpha G_0(T_K)).
\end{align}

This means that if we draw a random distribution from the DP and add
up the probability mass in a region $T \in \Theta$, then there will on
average be $G_0(T)$ mass in that region. The concentration parameter
plays the role of an inverse variance; for higher values of $\alpha$,
the random probability mass $G(T)$ will concentrate more tightly
around $G_0(T)$.


\cite{ferguson73} proved two properties of the Dirichlet process.  The
first property is that random distributions drawn from the Dirichlet
process are discrete.  They place their probability mass on a
countably infinite collection of points, called ``atoms,''
\begin{equation}
  \label{eq:discrete}
  G = \sum_{k=1}^{\infty} \pi_k \delta_{\theta^*_k}.
\end{equation}
In this equation, $\pi_k$ is the probability assigned to the $k$th
atom and $\theta^*_k$ is the location or value of that atom.  Further,
these atoms are drawn independently from the base distribution $G_0$.

The second property connects the Dirichlet process to the Chinese
restaurant process.  Consider a random distribution drawn from a DP
followed by repeated draws from that random distribution,
\begin{eqnarray}
  G &\sim& \textrm{DP}(\alpha, G_0) \\
  \theta_i &\sim& G \quad i \in \{1, \ldots, n\}.
\end{eqnarray}
\cite{ferguson73} examined the joint distribution of $\theta_{1:n}$,
which is obtained by marginalizing out the random distribution $G$,
\begin{equation}
  \label{eq:representation}
  p(\theta_1, \ldots, \theta_n \, | \, \alpha, G_0) =
  \int \left( \prod_{i=1}^{n} p(\theta_i \, | \, G) \right) dP(G \g
  \alpha, G_0).
\end{equation}
He showed that, under this joint distribution, the $\theta_i$ will
exhibit a \textit{clustering property}---they will share repeated
values with positive probability.  (Note that, for example, repeated
draws from a Gaussian do not exhibit this property.)  The structure of
shared values defines a partition of the integers from 1 to $n$, and
the distribution of this partition is a Chinese restaurant process
with parameter $\alpha$.  Finally, he showed that the unique values of $\theta_i$
shared among the variables are independent draws from $G_0$.

Note that this is another way to confirm that the DP assumes
exchangeability of $\theta_{1:n}$.  In the foundations of Bayesian
statistics, De Finetti's representation theorem \citep{finetti31} says
that an exchangeable collection of random variables can be represented
as a conditionally independent collection: first, draw a data
generating distribution from a prior over distributions; then draw
random variables independently from that data generating distribution.
This reasoning in Equation~\ref{eq:representation} shows that
$\theta_{1:n}$ are exchangeable.  (For a detailed discussion of De
Finetti's representation theorems, see~\cite{bernardo94}.)

\subsubsection*{Dirichlet process mixtures}

A DP mixture adds a third step to the model
above~\cite{Antoniak:1974},
\begin{eqnarray}
  G &\sim& \textrm{DP}(\alpha, G_0) \\
  \theta_i &\sim& G \\
  x_i &\sim& p(\cdot \g \theta_i).
\end{eqnarray}
Marginalizing out $G$ reveals that the DP mixture is equivalent to a
CRP mixture.  Good Gibbs sampling algorithms for DP mixtures are based
on this representation~\citep{escobar95,neal00}.

\begin{figure}
\centering
\includegraphics[width=0.8\textwidth]{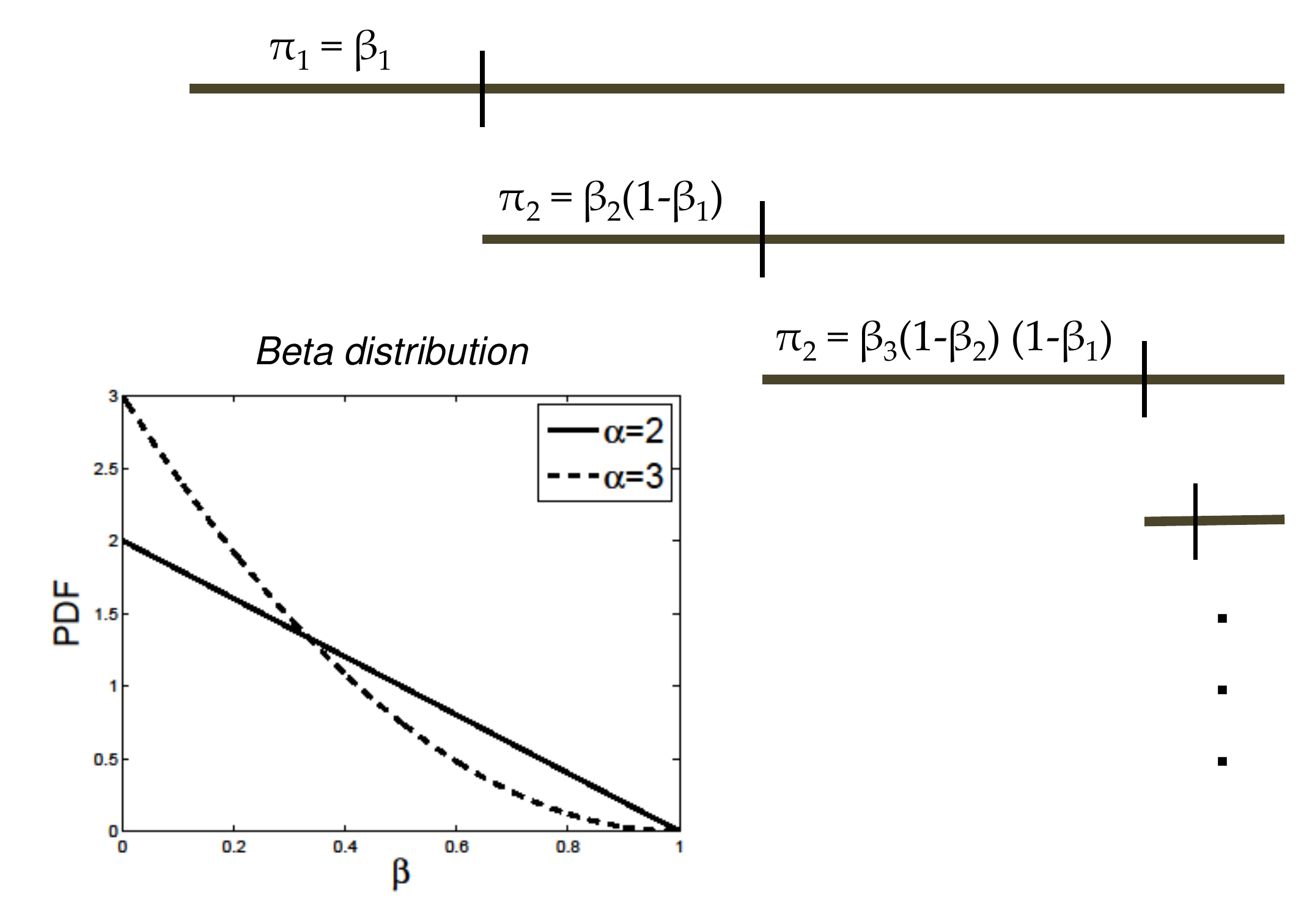}
\caption{\textbf{Stick-breaking construction}. Procedure for
  generating $\pi$ by breaking a stick of length 1 into
  segments. Inset shows the beta distribution from which $\beta_k$ is
  drawn, for different values of $\alpha$.}
	\label{fig:stick}
\end{figure}

\subsubsection*{The stick-breaking construction}

\cite{ferguson73} proved that the DP exists via its finite dimensional
distributions.  \citet{sethuraman94} provided a more constructive
definition based on the \textit{stick-breaking representation}.

Consider a stick with unit length.  We divide the stick into an
infinite number of segments $\pi_k$ by the following process.  First,
choose a beta random variable $\beta_1 \sim \textrm{beta}(1, \alpha)$
and break of $\beta_1$ of the stick.  For each remaining segment,
choose another beta distributed random variable, and break off that
proportion of the remainder of the stick.  This gives us an infinite
collection of weights $\pi_k$,
\begin{eqnarray}
  \beta_k &\sim& \mbox{Beta}(1,\alpha) \\
  \pi_k &=& \beta_k \prod_{j=1}^{k-1} (1-\beta_j) \quad k = 1, 2, 3, \ldots
\label{eq:gem}
\end{eqnarray}
Finally, we construct a random distribution using
Equation~\ref{eq:discrete}, where we take an infinite number of draws
from a base distribution $G_0$ and draw the weights as in
Equation~\ref{eq:gem}.  \cite{sethuraman94} showed that the
distribution of this random distribution is a $\textrm{DP}(\alpha,
G_0)$.

This representation of the Dirichlet process, and its corresponding
use in a Dirichlet process mixture, allows us to compute a variety of
functions of posterior DPs~\citep{Gelfand:2002} and is the basis for
the variational approach to approximate inference~\citep{blei06}.

\subsection*{\textbf{The beta process and Bernoulli process}}

Latent factor models admit a similar analysis \citep{thibaux07}. We
define the random measure $B$ as a set of weighted atoms:
\begin{align}
  B = \sum_{k=1}^K w_k \delta_{\theta_k},
  \label{eq:B}
\end{align}
where $w_k \in (0,1)$ and the atoms $\lbrace \theta_k \rbrace$ are drawn from a base measure $B_0$ on
$\Theta$. Note that in this case (in contrast to the DP), the sum of the weights does not sum to 1 (almost surely), which means that $B$ is not a probability
measure. Analogously to the DP, we can define a ``distribution on distributions'' for random measures with weights between 0 and 1---namely the beta process, which we denote by $B \sim \mbox{BP}(\alpha,B_0)$. Unlike the DP (which we could define in terms of Dirichlet-distributed marginals), the beta process cannot be defined in terms of beta-distributed marginals. A formal definition requires an excursion into the theory of completely random measures, which would take us beyond the scope of this appendix \citep[see][]{thibaux07}.

To build a latent factor model from the beta process, we define a new
random measure
\begin{align}
  X_n = \sum_{k=1}^K z_{nk} \delta_{\phi_k},
\end{align}
where $z_{nk} \sim \mbox{Bernoulli}(w_k)$. The random measure $X_n$ is
then said to be distributed according to a \emph{Bernoulli process}
with base measure $B$, written as $X_n \sim \mbox{BeP}(B)$. A draw
from a Bernoulli process places unit mass on atoms for which
$z_{nk}=1$; this defines which latent factors are ``on'' for the $n$th
observation. $N$ draws from the Bernoulli process yield an IBP-distributed binary matrix $\mathbf{Z}$, as shown by \citet{thibaux07}.

In the context of factor analysis, the factor loading matrix $\mathbf{G}$ is generated from this process by first drawing the
atoms and their weights from the beta process, and then constructing
each $\mathbf{G}$ by turning on a subset of these atoms according to
a draw from the Bernoulli process. Finally, observation $\mathbf{y}_n$ is generated according to Eq. \ref{eq:fa}.

\subsubsection*{Stick breaking construction of the beta process}

A ``double-use'' of the same breakpoints $\beta$ leads to a
stick-breaking construction of the beta process \citep{teh07}; see also
\citet{paisley10}. In this case, the weights correspond to the length
of the remaining stick, rather than the length of the segment that was
just broken off: $\pi_k = \prod_{j=1}^{k} (1-\beta_j)$.

\subsection*{The infinite limit of finite models}

In this section, we show BNP models can be derived by taking the
infinite limit of a corresponding finite-capacity model. For mixture
models, we assume that the class assignments $\mathbf{z}$ were
drawn from a multinomial distribution with parameters $\pi = \lbrace
\pi_1,\ldots,\pi_K \rbrace$, and place a symmetric Dirichlet
distribution with concentration parameter $\alpha$ on $\pi$. The
finite mixture model can be summarized as follows:
\begin{align}
  &\pi|\alpha \sim \mbox{Dir}(\alpha), \quad z_n|\pi \sim \pi \nonumber \\
  &\theta_k|G_0 \sim G_0, \quad \mathbf{y}_n|z_n,\theta \sim
  F(\theta_{z_n}).
\end{align}
When $K \rightarrow \infty$, this mixture converges to a Dirichlet
process mixture model \citep{neal92,rasmussen00,ishwaran02}.

To construct a finite latent factor model, we assume that each mask
variable is drawn from the following two-stage generative process:
\begin{align}
w_k|\alpha &\sim \mbox{Beta}(\alpha/K,1) \label{eq:beta} \\
z_{nk}|w_k &\sim \mbox{Bernoulli}(w_k).
\end{align}
Intuitively, this generative process corresponds to creating a bent
coin with bias $w_k$, and then flipping it $N$ times to determine
whether to activate factors $\lbrace z_{1k},\ldots,z_{Nk}\rbrace$.
\citet{griffiths05} showed that taking the limit of this model as $K
\rightarrow \infty$ yields the IBP latent factor model.

\section*{Appendix B: Software packages}

Below we present a table of several available software packages implementing the models presented in the main text.

\begin{center}
    \begin{tabular}{ | p{2cm} | p{2cm} | p{2cm} | p{3cm} | p{5.5cm} |}
    \hline
    \textbf{Model} & \textbf{Algorithm} & \textbf{Language} & \textbf{Author} & \textbf{Link} \\ \hline
    CRP mixture model & MCMC & Matlab & Jacob Eisenstein & \url{http://people.csail.mit.edu/jacobe/software.html} \\ \hline
    CRP mixture model & MCMC & R & Matthew Shotwell & \url{http://people.csail.mit.edu/jacobe/software.html} \\ \hline
    CRP mixture model & Variational & Matlab & Kenichi Kurihara & \url{http://sites.google.com/site/kenichikurihara/academic-software} \\ \hline
    IBP latent factor model & MCMC & Matlab & David Knowles & \url{http://mlg.eng.cam.ac.uk/dave} \\ \hline
    IBP latent factor model & Variational & Matlab & Finale Doshi-Velez & \url{http://people.csail.mit.edu/finale/new-wiki/doku.php?id=publications_posters_presentations_code} \\ \hline
    \end{tabular}
\end{center}

\newpage

\bibliographystyle{apalike}
\bibliography{bib}

\end{document}